\documentclass{article}
\usepackage{mlspconf,amsmath,graphicx}

\usepackage[utf8]{inputenc} 
\usepackage[T1]{fontenc}    
\usepackage{hyperref}       
\usepackage{url}            
\usepackage{booktabs}       
\usepackage{amsfonts}       
\usepackage{amsmath}       
\usepackage{nicefrac}       
\usepackage{microtype}      

\usepackage{graphicx}
\graphicspath{{figs/}}
\usepackage{caption}
\usepackage{subcaption}
\usepackage[export]{adjustbox}
\usepackage{wrapfig}
\usepackage{xcolor}
\usepackage{amsthm}
\usepackage{stmaryrd}
\usepackage{mathtools}
\usepackage{shortcuts}
\usepackage[percent]{overpic}

\newcommand{\abs}{\text{abs}}

\copyrightnotice{
\begin{minipage}[t]{\textwidth}\
  {\copyright}2019 IEEE. Personal use of this material is permitted. Permission from IEEE must be obtained for all other uses, in any current or future media, including reprinting/republishing this material for advertising or promotional purposes, creating new collective works, for resale or redistribution to servers or lists, or reuse of any copyrighted component of this work in other works.
\end{minipage}}

\title{Self-supervised representation learning from electroencephalography signals}
  
\begin{document}
%
\name{
    \begin{minipage}{\linewidth}
        \centering
        Hubert Banville$^{1,2}$ \qquad Isabela Albuquerque$^{3}$ \qquad Aapo Hyv\"arinen$^{1,4}$ \\
        Graeme Moffat$^{2}$ \qquad Denis-Alexander Engemann$^{1}$ \qquad Alexandre Gramfort$^{1}$
    \end{minipage}
}

\address{$^{1}$ Inria, Universit\'e Paris-Saclay, Paris, France\\
    $^{2}$ InteraXon Inc., Toronto, Canada\\
    $^{3}$ INRS-EMT, Universit\'e du Qu\'ebec, Montr\'eal, Canada \\
    $^{4}$ Dept. of CS and HIIT, University of Helsinki, Finland}

\maketitle

\begin{abstract}
The supervised learning paradigm is limited by the cost - and sometimes the impracticality - of data collection and labeling in multiple domains.
Self-supervised learning, a paradigm which exploits the structure of unlabeled data to create learning problems that can be solved with standard supervised approaches, has shown great promise as a pretraining or feature learning approach in fields like computer vision and time series processing.
In this work, we present self-supervision strategies that can be used to learn informative representations from multivariate time series.
One successful approach relies on predicting whether time windows are sampled from the same temporal context or not.
As demonstrated on a clinically relevant task (sleep scoring) and with two electroencephalography datasets, our approach outperforms a purely supervised approach in low data regimes, while capturing important physiological information without any access to labels.
\end{abstract}

\section{Introduction}

The impressive success of deep learning in various domains can in large part be explained by the availability of large labeled datasets, such as COCO \cite{lin2014microsoft} for object recognition or LibriSpeech \cite{panayotov2015librispeech} for speech recognition.
While such annotated datasets enable the use of supervised learning methods for which experimentation and validation are well understood, they must first be labeled - a generally costly and time consuming process if possible at all.
Indeed, labeling can be particularly challenging for certain types of data that are highly complex or noisy, resulting in poor quality human annotations at best.
When data are available in large amounts but labels are missing, the classical approach is to rely on unsupervised statistical models such as clustering or latent factor models. 
However, choosing the unsupervised method remains a challenge as the right criterion may not be obvious.

Self-supervised learning (SSL) is a recently developed area of research that provides a compelling approach to making use of large unlabeled datasets.
With SSL, the structure of the data is used to turn an unsupervised learning problem into a supervised learning problem, called a ``pretext task''~\cite{jing2019self}.
The representation learned on the self-supervised pretext task can then be reused on a supervised downstream task, potentially greatly reducing the number of labeled examples required.


Learning paradigms that exploit the structure of unlabeled data have been used in various domains such as computer vision (CV) (see~\cite{jing2019self} for a recent review) and time series analysis.
%
In~\cite{doersch2015unsupervised}, models were trained to predict the relative position of image patches and then fine-tuned on various downstream supervised tasks.
With this approach, an R-CNN pretrained on the target dataset (Pascal VOC) with SSL achieved similar performance as a network pretrained on ImageNet labels.
SSL has also been used to learn visual representations from videos.
In the frame contiguity prediction task of Misra et al.~\cite{misra2016shuffle}, a siamese network was trained to predict whether tuples of three video frames were in the right temporal order.
This approach led to improved performance over purely supervised models when SSL models were used to initialize network weights.
SSL has also been applied to time series data.
For instance, autoregressive models based on an encoding principle achieved promising performances on speech data.
In \cite{oord2018representation}, an encoder first projects windowed data into a latent space; an autoregressive model then summarizes the previous encodings into a contextual latent representation.
Finally, the model is trained to predict the next $N$ encodings based on the current context latent representation.
The authors showed improved downstream performance on a speaker identification task as compared to other SSL approaches.
%

A general and theoretically grounded approach to SSL was recently formalized by Hyv\"arinen et al. \cite{hyva17aistats, hyvarinen2019nonlinear} from the perspective of nonlinear independent components analysis.
Under that generalized framework, SSL tasks are constructed by using an auxiliary variable $\mathbf{u}$ (e.g., the time index, the index of a segment or the history of the data) to train a contrastive classifier.
This classifier learns to predict whether a sample $\mathbf{x}$ is paired with its correct auxiliary variable $\mathbf{u}$ or a perturbed (random) one $\mathbf{u^*}$.
Most of the previously introduced SSL tasks can be framed under this formulation.

Although most applications to date have focused on tasks for which plentiful annotated data are already available (object detection, language modelling, etc.), SSL could prove particularly useful in fields where low labeled data regimes are common such as physiological signal analysis.
Indeed, labels for biosignals such as electroencephalography (EEG) are often difficult to obtain as they require extensive expert knowledge.
For instance, sleep staging, i.e., the task of identifying the different sleep stages in recordings of sleep, requires trained technicians to manually annotate hours of data \cite{younes2017case}.
For epilepsy and other pathological conditions, recordings must be annotated by neurologists and other medical professionals.
Learning useful representations automatically from unlabeled biosignals could therefore drastically reduce the cost and time required to process such signals.

In this paper, we propose self-supervised strategies to learn end-to-end features from unlabeled time series such as EEG.
We introduce two temporal contrastive learning tasks that we refer to as ``relative positioning'' and ``temporal shuffling''.
Experimentally, we show that these contrastive learning tasks based on predicting whether time windows are close in time can be used to learn EEG features that capture multiple components of the structure underlying the data.
We demonstrate that these features, when reused on a downstream sleep staging task, outperform traditional unsupervised and purely supervised approaches, specifically in low-data regimes.
Moreover, we show that models trained with these approaches learn physiologically meaningful representations.

The rest of the paper is structured as follows. 
Section~\ref{sec:learning} describes the SSL tasks and learning problems considered. 
Section~\ref{sec:methods} describes the data, the neural architectures and results on three experiments. 
Lastly, in Section~\ref{sec:discussion}, we discuss the results and conclude.

\section{Self-supervised learning with temporal contrastive tasks}
\label{sec:learning}

\paragraph*{Notation}
We denote by $\intset{q}$ the set $\{1, \ldots, q\}$ for any integer $q \in \bbN$.
We denote by 
$i \in \intset{N}$ the i$^{th}$ sample in the training set of cardinality $N$.
The index $t$ refers to time indices in the input multivariate time series
$S \in \bbR^{M \times C}$, where $M$ is the number of times samples and
$C$ is the dimension of samples (\eg channels). We assume for simplicity
that each $S$ has the same size. We denote by $y \in \{-1, 1\}$
a binary label used in the learning task.

\subsection{Deep time contrastive learning with relative positioning and temporal shuffling as pretext tasks}

To produce labeled samples from the multivariate time series $S$, we propose to sample pairs of time windows $(x_{t}, x_{t'})$ where each window $x_t$, $x_{t'}$ is in $\bbR^{T \times C}$, and $T$ is the duration of each window.
The first window $x_{t}$ is referred to as the ``anchor window''.
Our assumption is that an appropriate representation of the data should evolve slowly over time suggesting that time windows close in time should share the same label.
Given $\tau_{pos} \in \bbN$, which controls the duration of the positive context, and $\tau_{neg} \in \bbN$, which corresponds to the negative context around each window $x_i$, we sample $N$ labeled pairs:
\begin{equation*}
\mathcal{Z}_N = \{ ((x_{t_i}, x_{t_i'}), y_i) | i \in \intset{N}, (t_i, t'_i) \in \cT \times \cT, y_i \in \cY\},
\end{equation*}
where $\cY = \{-1, 1\}$, $\cT = \{(t,t') \in \intset{M - T + 1}^2 /\ |t - t'| \leq \tau_{pos} \textrm{\ or\ } |t - t'| > \tau_{neg}\}$. Here $y_i \in \cY$ is specified by the positive or negative contexts parameters:
\begin{equation}   
y_{i} = 
     \begin{cases}
       \text{1,} &\quad\text{if}\;|t_i-t_i'|\leq\tau_{pos}\\
       \text{-1,} &\quad\text{if}\;|t_i-t_i'|>\tau_{neg}\\
     \end{cases}.
\end{equation}
We ignore window pairs where $x_{t'}$ falls outside of the positive and negative contexts of the anchor window $x_t$.
In other words, the label indicates whether two time windows are closer together than $\tau_{pos}$ or farther apart than $\tau_{neg}$ in time.
We call this pretext task ``relative positioning'' (RP).

We also introduce a variation of the RP task that we call ``temporal shuffling'' (TS).
In the TS task, we sample a third window $x_{t''}$ from the positive context of $x_t$ and use it to provide an additional point of reference to compare $x_{t'}$ with.
From this perspective, the label $y_i$ now indicates whether the three windows are temporally ordered ($t < t' < t''$) or whether the windows have been shuffled ($t' < t$ or $t' > t''$), similar to \cite{misra2016shuffle}

In order to learn end-to-end how to discriminate tuples of time windows based on their relative position or order, we introduce a feature extractor $h: \bbR^{T \times C} \rightarrow \bbR^D$ with parameters $\Theta$ which maps a window $x$ to its representation on the feature space.
A contrastive module is then used to aggregate the feature representations of each window.
For the RP task, $g_{RP}: \bbR^D \times \bbR^D \rightarrow \bbR^D$ combines representations from pairs of windows, for example by computing an elementwise absolute difference, denoted by the $\abs$ operator: $g_{RP}(h(x), h(x')) = \abs(h(x) - h(x')) \in \bbR^D$.
For TS, $g_{TS}: \bbR^D \times \bbR^D \times \bbR^D \rightarrow \bbR^{2D}$ can be implemented by concatenating the absolute differences $g_{TS}(h(x), h(x'), h(x'')) = (\abs(h(x) -h(x')), \abs(h(x')-h(x''))) \in \bbR^{2D}$.

Finally, a linear context discriminative model with coefficients $w \in \bbR^D$, or $\in \bbR^{2D}$, and bias term $w_0 \in \bbR$ is responsible for predicting the associated target $y$.
Using the binary logistic loss on the predictions of $g$ (either the RP or TS variant), we can write a joint loss function $\mathcal{L}(\Theta,w,w_0)$ as
\begin{multline}
\mathcal{L}(\Theta,w,w_0) = \\
  \smashoperator{\sum_{(x_t, x_{t'}, y) \in \mathcal{Z}_N}}
    \log(1 + \exp(-y [w^\top g(h(x_t), h(x_{t'})) + w_0])),
\end{multline}
which we assume fully differentiable with respect to the parameters $(\Theta,w,w_0)$. Given
the convention used for $y$, the predicted target is the sign of $w^\top g(h(x_t), h(x_{t'})) + w_0$.

Both the RP and TS models can be seen as siamese neural networks with two or three subnetworks, respectively.





\section{Application to EEG sleep data}
\label{sec:methods}

\subsection{Data and preprocessing steps}

We conduct our experiments on two openly available datasets of EEG sleep data (see Table~\ref{tab:datasets}).
The Physionet Sleep EDF expanded dataset \cite{kemp2000analysis, goldberger2000physiobank} contains 153 sleep recordings from 83 healthy subjects (age 25 to 101).
EEG channels Fpz-Cz and Pz-Oz were recorded at 100~Hz. 
Windows of 30~s were labeled by trained sleep technicians following the R\&K definition of sleep stages, however we combined sleep stages 3 and 4 to follow the AASM manual~\cite{berry2012aasm}. 
This yields five labels: W (wake), N1, N2 and N3 for different levels of sleep (N3 are the deep sleep periods), and R (rapid eye movements, REM).

The second dataset is the MASS dataset session 3 \cite{o2014montreal}. 
It contains a single whole-night sleep recording for 62 healthy subjects from 20 to 69 years old.
A total of 20 EEG channels sampled at 256~Hz were recorded following the standard 10-20 system and using a linked-ear reference.
The 30-s windows were labeled following the five stages of the AASM manual. 

\begin{table}[h]
\centering
\caption{Description of the two datasets used in this study, Sleep EDF and MASS, as well as the available number of samples per class in each dataset.}
\label{tab:datasets}
\begin{tabular}{@{}lll@{}}
\toprule
                   & Sleep EDF \cite{kemp2000analysis} & MASS (SS3) \cite{o2014montreal} \\ \midrule
W                  &    59,982    & 6,131     \\
N1                 &    21,522    & 4,709     \\
N2                 &    69,132    & 28,920    \\
N3                 &    13,039    & 7,362     \\
R                  &    25,835    & 10,323     \\ \midrule
\# subjects        & 83           &  62    \\
\# recordings      & 153          &  62    \\
Sampling frequency & 100 Hz       &  256 Hz    \\
\# EEG channels    & 2 bipolar    & 20     \\
Reference          & -            & Linked ears    \\ \bottomrule
\end{tabular}
\end{table}


For both datasets, the raw EEG channels were filtered using a 30~Hz 4th-order FIR lowpass filter.
MASS recordings were downsampled to 128~Hz, and channels Fz, Cz and Oz were extracted to reduce input dimensionality.
Non-overlapping windows of 30~s were extracted, yielding windows of size $T=2000$ and $C=2$ on Sleep EDF, and $T=3840$ with $C=3$ on MASS.
The windows were normalized so that channels had mean 0 and standard deviation 1.

A total of 2000 anchor windows were sampled uniformly within each recording.
For each anchor window, three positive and three negative tuples were sampled.
On Sleep EDF, subjects 40 to 82 were used for training, subjects 0 to 19 for validation, and subjects 20 to 39 for testing\footnote{Missing subjects and sessions were: subject 13 session 2, subject 36 session 1, subject 52 session 1, as well as subjects 39, 68, 69, 78 and 79.}.
On MASS, subjects 1 to 41 were used for training, 42 to 52 for validation, and 52 to 62 for testing.\footnote{Subjects 43 and 49 were not available.}
This yielded a total of 512,622, 267,630 and 342,300 pairs in the training, validation and test splits for Sleep EDF, and 237,882, 52,152 and 73,650 pairs for MASS.

\subsection{Model architecture}

For the feature extractor $h$, we adapt a previously published architecture shown to perform well on sleep staging~\cite{chambon2018deep}.
For an input size $(C, T, 1)$ where $C$ is the number of EEG channels and $T$ is the number of time points in a window, the CNN is defined as: 
$conv(C \times 1, C) \rightarrow permute(2, 1, 0) \rightarrow conv(1 \times k, 8) \rightarrow ReLU \rightarrow maxpool(1 \times m) \rightarrow conv(1 \times k, 8) \rightarrow ReLU \rightarrow maxpool(1 \times m) \rightarrow flatten \rightarrow dropout(50\%) \rightarrow linear(C \times (T //k //k) \times 8, D)$.
We set the filter size $k$ and maxpooling size $m$ to 50 and 13 for Sleep EDF, and to 64 and 16 for MASS. In both cases the embedding dimension is $D=100$.
This results in 55,545 trainable parameters for Sleep EDF and 67,173 for MASS.


The Adam optimizer \cite{kingma2014adam} with $\beta_1$=0.9, $\beta_2$=0.999 and learning rate 0.001 is used, while the batch size is 256. Training runs up to 300 epochs, or until the validation loss does not decrease anymore for a period of at least 30 epochs.
Dropout is applied to fully-connected layers at a rate of 50\%.

\subsection{Compared models}
We compare the performance of our model trained on the SSL tasks to three neural network baselines: 1) random initialization (rand init), 2) convolutional autoencoder (AE) \cite{masci2011stacked} and 3) purely supervised learning.
The AE model uses $h$ as the encoder and a four-layer convolutional decoder, along with mean squared error as the reconstruction loss.
In the case of the purely supervised model, an additional softmax layer is added to the feature extractor $h$ to classify labeled epochs into one of five sleep stages.
We use \textit{pytorch} \cite{paszke2017automatic} and \textit{scikit-learn} \cite{pedregosa2011scikit} to build and train all models.

As an additional point of comparison, we also extract human-engineered EEG features \cite{aboalayon2016sleep}:
mean, variance, skewness, kurtosis, standard deviation, frequency log-power bands between $(0.5, 4, 8, 13, 30, 49)$~Hz as well as all their possible ratios, peak-to-peak amplitude, Hurst exponent, approximate entropy and Hjorth complexity.
This results in 34 features per EEG channel, which are concatenated into a single vector.

To account for class imbalance, we use balanced accuracy (bal acc), defined as the average per-class recall, to evaluate model performance on the downstream task.
Moreover, during training, the loss is weighted to account for class imbalance.

\subsection{Experiments}

We present three experiments designed to evaluate the SSL tasks in the context of EEG classification, and demonstrate their usefulness for learning from unlabeled data in clinically relevant scenarios.
In the first experiment, in order to validate to the SSL tasks, we analyze the performance of the CNN for different SSL hyperparameter values as well as their impact on the sleep staging downstream task.
In the second experiment, we probe the ability of the SSL tasks to improve prediction performance with limited annotated data.
Finally, in the third experiment, we explore the features learned through SSL and study their physiological relevance.

\subsubsection{Experiment 1: SSL models learn representations of EEG signals and facilitate sleep staging}

We first evaluate the ability of the CNN architecture to learn on the SSL tasks (see Table~\ref{tab:hyperparams}).
We train the feature extractor $h$ on the entire training set using the RP and TS tasks with three sets of hyperparameters $\tau_{pos}$ and $\tau_{neg}$.
Once $h$ is trained, we project the labeled samples into the networks' respective feature space and then train multinomial linear logistic regression models on each set of features to predict sleep stages.

On MASS, $\tau_{pos}=2$, $\tau_{neg}=2$ (in minutes) and $\tau_{pos}=4$, $\tau_{neg}=15$ both led to similar performances on the SSL and downstream tasks.
Making the task harder by using a large positive context ($\tau_{pos}=120$) however led to lower performance on both SSL and sleep staging tasks.
The same conclusion was reached on Sleep EDF (results not shown).
We decided to use $\tau_{pos}=4$ and $\tau_{neg}=15$ in our experiments, as this also increases the number of windows that can be sampled from the positive context.

While these results are a few points below those of a linear classifier trained on common handcrafted EEG features (79.43$\%$), they are slightly better than full supervision (72.51$\%$), showing our approach achieves comparable performance to standard approaches but without labels or expert knowledge.
SSL performance was similar to full supervision or common EEG features on Sleep EDF as well.

\begin{table}[h]
\centering
\caption{Test balanced accuracy obtained on the SSL tasks (bal acc$_{SSL}$) and on the sleep staging task (bal acc$_{staging}$) for different sets of hyperparameters $\tau_{pos}$ and $\tau_{neg}$ (in minutes). Results obtained on MASS.}
\label{tab:hyperparams}
\begin{tabular}{@{}lrrll@{}}
\toprule
   & $\tau_{pos}$ & $\tau_{neg}$ & bal acc$_{SSL}$ & bal acc$_{staging}$ \\
\midrule
RP & 2                  & 2                  &  \textbf{79.49}                &           75.73           \\
   & 4                 & 15                 & 78.60                 &     \textbf{76.66}                       \\
   & 120                & 120                &  56.30                &      65.71                     \\
\midrule
TS & 2                  & 2                  &   81.42               &       \textbf{75.90}                    \\
   & 4                 & 15                 &   \textbf{82.12}               &      75.37                      \\
   & 120                & 120                &    66.59              &      66.66                      \\
\midrule
EEG features & - & - & - & 79.43 \\
Fully supervised & - & - & - & 72.51 \\
\bottomrule
\end{tabular}
\end{table}

\subsubsection{Experiment 2: SSL enables sleep staging with limited annotated data}

\begin{figure}[htb]
\begin{minipage}[b]{1.0\linewidth}
  \centering
  \begin{overpic}[width=1\textwidth]{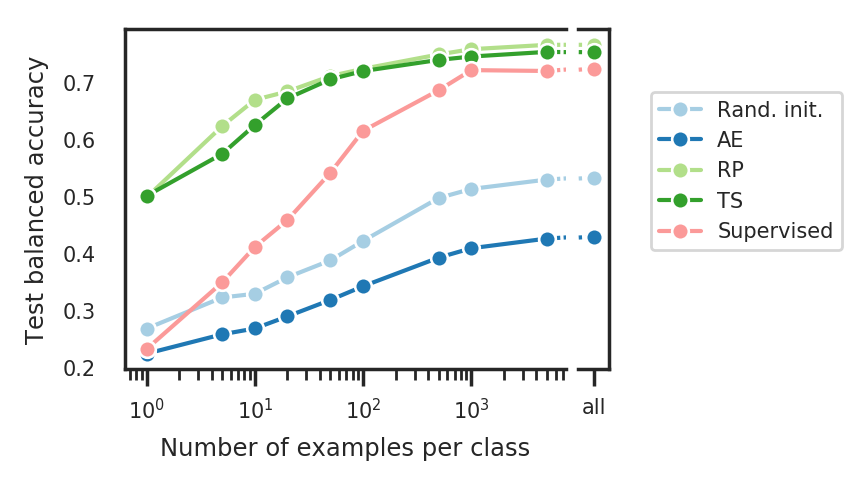}
  \put (1,1) {\large{A)}}
  \end{overpic}
\end{minipage}
\begin{minipage}[b]{1.0\linewidth}
  \centering
  \begin{overpic}[width=1\textwidth]{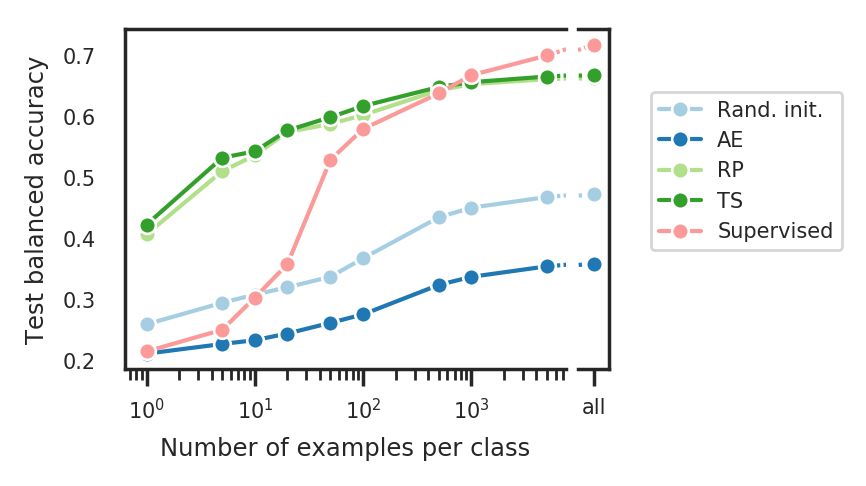}
  \put (1,1) {\large{B)}}
  \end{overpic}
\end{minipage}
\caption{Impact of number of labeled examples per task on sleep staging performance for feature extractors trained with an autoencoder (AE), the relative positioning (RP) and the temporal shuffling (TS) tasks, a fully supervised model, and a randomly initialized model, \textbf{(A)} on MASS and \textbf{(B)} on Sleep EDF. ``All'' means all available training examples were used. While higher numbers of labeled examples lead to better performance, SSL models achieve much higher performance as supervised models when few examples are available.}
\label{fig:exp2_quantity_data}
\end{figure}

Next, in order to assess whether models trained with SSL can learn informative features, we compare their performance to the performance of various baselines, and explore the effect of varying the quantity of labeled data (see Fig.~\ref{fig:exp2_quantity_data}).

The feature extractors $h$ are trained using the different approaches (AE, RP and TS on unlabeled data; full supervision on labeled data) and then used to obtain features.
We also extract features using models with randomly initialized weights, i.e., that have not been trained.
The sleep staging performance is finally evaluated using logistic regression models.

On MASS (Fig.\ref{fig:exp2_quantity_data}-A), the SSL features outperformed the purely supervised model for all data regimes, with more than 25 points difference when a single example per class was available.
Across most data regimes, RP was found to outperform TS by a fraction of a percent.
Finally, both AE and the randomly initialized model led to much lower performance, albeit over chance level ($\sim20\%$).
Similar results were obtained on Sleep EDF (Fig.\ref{fig:exp2_quantity_data}-B), although the purely supervised model outperformed SSL features above 500 examples per class.
TS also led to slightly higher performance than RP.

The model pretrained with an autoencoder obtained very low performance because the reconstruction task, which uses a mean squared error loss, encourages the model to focus on the input signal's low frequencies.
Indeed, these frequencies have higher power than high frequencies in biological signals like EEG.
While the autoencoder misses spectral information important for the sleep staging task, the SSL models show relatively high performance.

\subsubsection{Experiment 3: SSL models learn physiologically meaningful features}

To further explore the features learned with SSL, we project the 100-dimension embeddings obtained on the labeled Sleep EDF dataset to two dimensions using UMAP \cite{mcinnes2018umap}.
We use Sleep EDF as it contains subject metadata such as age.

\begin{figure}[htb]
\begin{minipage}[b]{0.90\linewidth}
  \centering
  \begin{overpic}[width=1\textwidth]{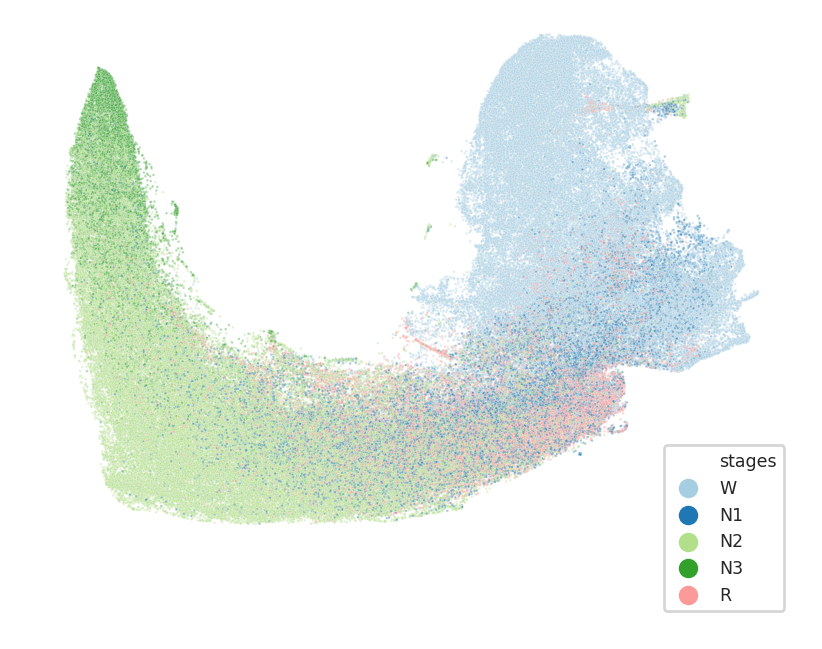}
  \put (1,1) {\large{A)}}
  \end{overpic}
\end{minipage}
\begin{minipage}[b]{0.90\linewidth}
  \centering
  \begin{overpic}[width=1\textwidth]{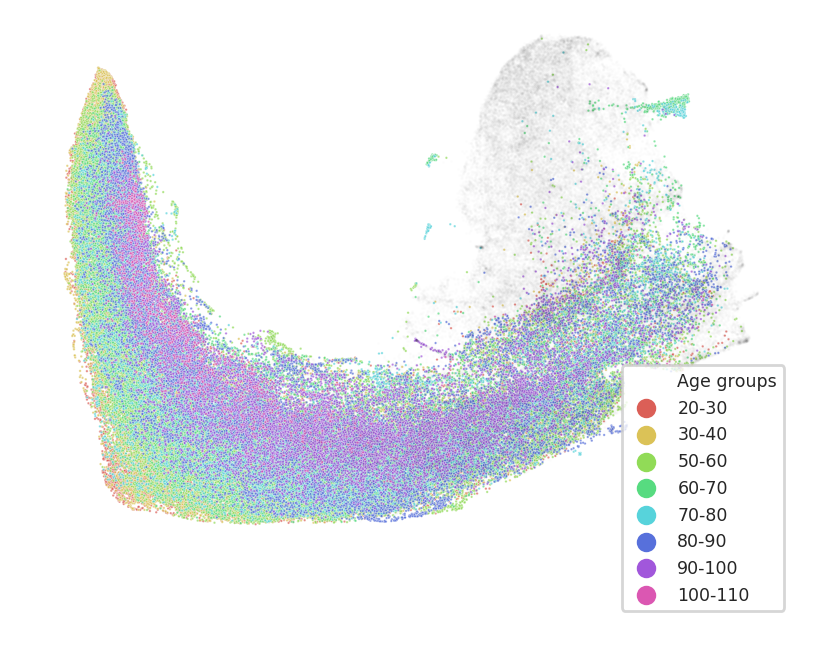}
  \put (1,1) {\large{B)}}
  \end{overpic}
\end{minipage}
\caption{UMAP visualization of temporal shuffling (TS) features on the whole Sleep EDF dataset. Each point corresponds to the features extracted from a 30-s window of EEG. \textbf{(A)} Samples are color-coded by sleep stage. \textbf{(B)} Samples for stages N1, N2 and N3 are color-coded by age group while other stages are in grey. As seen from their gradient-like structure, the features encode physiologically relevant information although no labels were available during training.}
\label{fig:exp3_embs}
\end{figure}

In Fig.~\ref{fig:exp3_embs}-A, we notice the emergence of a distinct structure that closely follows the different sleep stages.
Indeed, as seen by color-coding samples using their labels, clear groups  emerge that not only correspond to the labeled sleep stages, but that are also sequentially arranged: starting from the right of the figure and moving to the left, we can draw a trajectory that passes through W, N1, N2 and N3 sequentially.
Stage R, finally, overlaps with W and N1.

Moreover, in Fig.~\ref{fig:exp3_embs}-B, one can observe that the embedding encodes age-related information.
Samples from younger subjects occupy the left outer part of the cloud of points, while samples from older subjects are found in the inner part of the U-shaped structure.
This phenomenon is visible in stages N1, N2 and N3 but not in W and R, where no apparent age-dependent structure is visible.
This might be explained by the prevalence of sleep spindles, major features used to identify N2 and N3, which are known to change with age~\cite{purcell2017characterizing}.


\section{Discussion}
\label{sec:discussion}

We introduced two self-supervised learning tasks, relative positioning and temporal shuffling, which we used to learn representations from electroencephalography (EEG) multivariate time series.
Our approach achieves similar performance as supervised approaches when tested on a clinically relevant sleep staging task, and largely outperforms a purely supervised approach in lower data regimes.
The representation learned with these tasks also encodes physiologically interesting structure such as sleep stages and age, demonstrating their potential to uncover meaningful latent structure in unlabeled data.

While both the SSL tasks proposed were shown to be useful for unsupervised training of feature extractors and achieved very similar performance, RP required fewer computations as its implementation uses only two siamese subnetworks, instead of three.
It might therefore be a better choice thanks to its relative simplicity.
The quality of the representation obtained with the RP task as well as its training efficiency could be further improved by 1) modifying the contrastive module $g$ to compute other aggregates of the features (e.g., sum or dot product) and 2) by using mining strategies, by which tuples are not sampled uniformly but rather based on a predefined criterion.
Indeed, the number of training examples (i.e., tuples of two or three windows) available to the RP and TS tasks can be made very high as the number of possible tuples increases exponentially with the number of available windows, which also increases training time.
Mining most useful examples could therefore speed up model training.

By reducing the number of labeled data required to reach high performance, the proposed SSL tasks are promising alternatives to an expensive and time-consuming labeling process, as well as expert handcrafting of task-specific features.
Future work will focus on evaluating the usefulness of the SSL task for other types of neural time series recordings, and on assessing the impact of architecture variations and mining strategies.



\bibliographystyle{IEEEbib}
\bibliography{main.bib}  

\end{document}